\documentclass[letterpaper]{article}
\usepackage{aaai}
\usepackage{times}
\usepackage{helvet}
\usepackage{courier}
\usepackage{multirow}
\usepackage{graphicx} 
\usepackage{diagbox}
\usepackage{lipsum} 
\usepackage{bm}
\usepackage[ruled,vlined,onelanguage]{algorithm2e}

\begin{document}
\title{Constructing a Highlight classifier with an Attention based LSTM Neural Network}
\author{Marius A.~Radu\\
FocusVision Worldwide Inc\\
Innovation Dept.\\
Stamford, CT 06902 \\
   \And
Michael H.~Kuehne\thanks{Direct all correspondence to mkuehne@focusvision.com.} \\
FocusVision Worldwide Inc\\
Data Innovation Dept.\\
Stamford, CT 06902 \\
}
\maketitle
\begin{abstract}
\begin{quote}
Data is being produced in larger quantities than ever before in human history. It's only natural to expect a rise in demand for technology that aids humans in sifting through and analyzing this inexhaustible supply of information. This need exists in the market research industry, where large amounts of consumer  research data is collected through video recordings. At present, the standard method for analyzing video data is human labor. Market researchers manually review the vast majority of consumer research video in order to identify relevant portions - \textit{highlights}. The industry state of the art turnaround ratio is $\sim$ 2.2 - for every hour of video content 2.2 hours of manpower are required. In this study we present a novel approach for NLP-based highlight identification and extraction based on a supervised learning model that aids market researchers in sifting through their data. Our approach hinges on a manually curated user-generated highlight clips constructed from long and short-form video data. The problem is best suited for an NLP approach due to the availability of video transcription. We evaluate multiple classes of models, from gradient boosting to recurrent neural networks, and compare their performance in extraction and identification of highlights. The best performing models are then evaluated using four sampling methods designed to analyze documents much larger than the maximum input length of the classifiers. We report very high performances for the standalone classifiers, ROC AUC scores in the range 0.93-0.94, but observe a significant drop in effectiveness when evaluated on large documents. Based on our results we suggest combinations of models/sampling algorithms for various use cases.

\end{quote}
\end{abstract}

\section{Introduction}

The field of data science has made tremendous progress over the last few decades in developing novel natural language processing (NLP) techniques which today have become a standard addition to any data scientist's repertoire: text summarization, information extraction, sentiment analysis to name a few. In this study we investigate a new approach to text abstraction - highlight analysis, and present a supervised learning algorithm to classify text. The study takes place in the context of the market research industry, our data-set consisting of interviews and focus groups. \cite{market_research_review}

\begin{figure}[ht]
\centering
{
\includegraphics[width=1\linewidth]{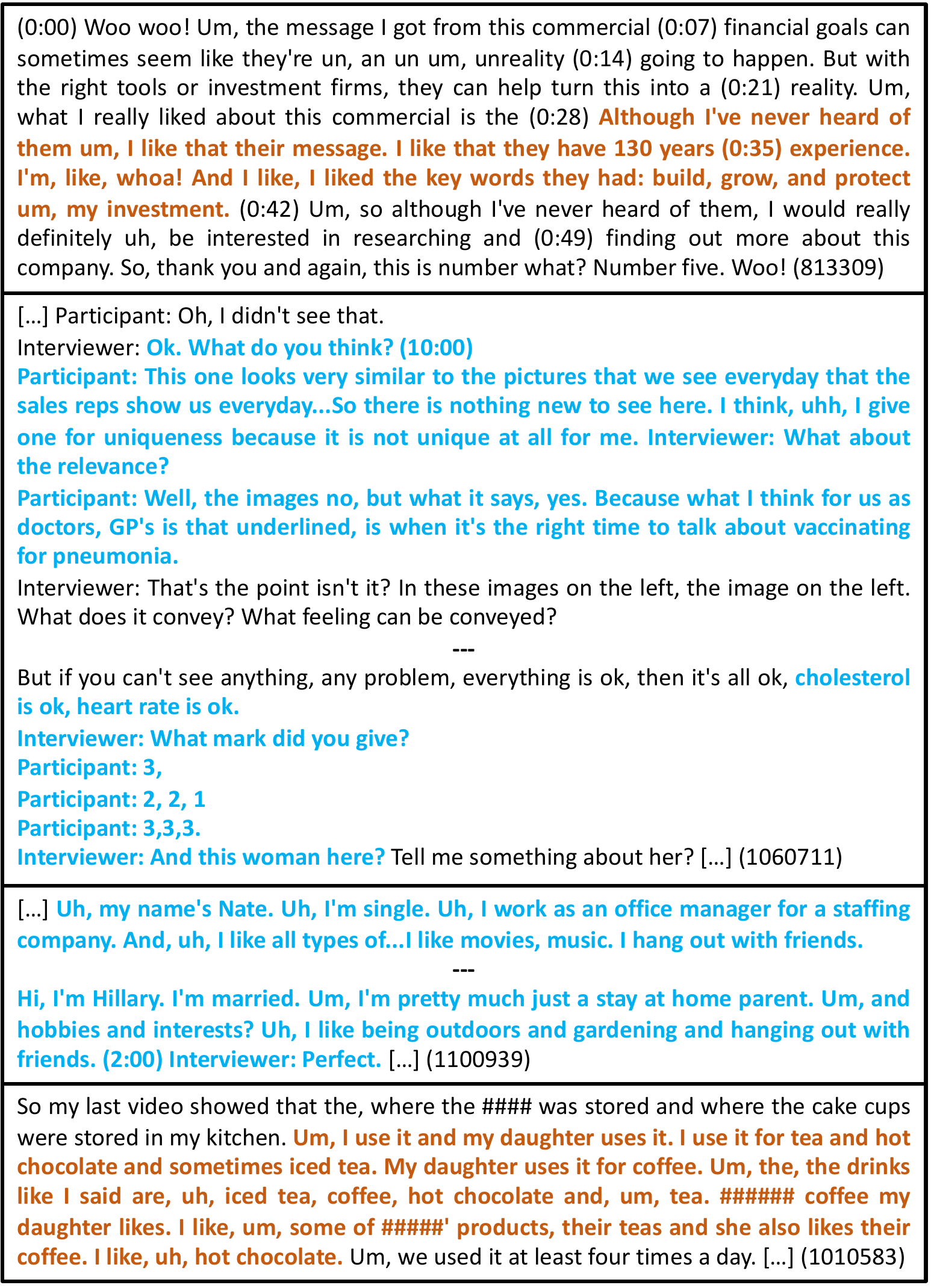}
}
\caption{Examples of user generated clips originating from one on one interviews (orange) and focus groups (blue). Transcript lengths can vary dramatically: 1, 30, 42, 4 minutes (top to bottom) as well as semantic content of clip. A very common phenomenon is for clips to begin(end) before(after) the apparent region of interest.}
\label{fig1}
\end{figure}

Market research is a field that pulls from a wide array of disciplines such as anthropology, statistics and organizational theory. Market researchers regularly collect customer data and they typically have access to volumes of organization data. Their primary job function is to analyze data, most of it being explicit (record based) knowledge and apply their tacit (contextual) knowledge to support decisions for their business.

Two primary data collection methods of the market research industry are focus groups and 1:1 web based video interviews. A focus group is a gathering of targeted individuals that participate in a planned discussion designed to extract consumer perceptions about a particular product, topic, or area of interest. Typically, the environment is designed with receptiveness in mind, encouraging unbiased sharing of opinions. Members of a group are often encouraged to interact and influence each other during the discussion for sharing and consideration of ideas and perspectives. The 1:1 interviews are executed similarly to focus groups, the main difference being the lack of group dynamics found in focus groups. Like focus groups, 1:1 interviews will be answering a variety of questions regarding a product, topic, or area of interest. It is standard practice to record these interactions, analyze them to use as decision support. In recent years market research video content has become very abundant. 

Manual highlight extraction is currently the industry standard, despite it being a time intensive and tiresome task. Inquiries into the costs and turnaround ratios of some of the industry leaders yielded a best estimate around 2.2 hours - 2.2 hours of manpower for every 1 hour of video content - along with the hefty price tag of $\$800$ per hour of video. An automated highlight extraction tool is therefore exceedingly desirable in the market research industry. The purpose of such a tool can be very broad, as it does not have to be limited to the task of generating highlight clips, but can serve as a generalized highlight classification model/filter, similar in scope to that of sentiment classification. 

In the field of market research, and in general commercial tasks, there is an ongoing search for a high-utility metric - one that yields statistically significant actionable information for clients. From a client's point of view, the best metric is one which demonstrates a direct correlation with the effect of public discourse on e.g. a brand or corporate reputation. Sentiment analysis would seem like the best candidate, however, it has been repeatedly called into question. \cite{sentiment2} It has been argued that without an underlying knowledge model, sentiment analysis alone shows little sustainable impact and low predictive power. \cite{sentiment1,sentiment3} In the broadest sense, the scope of this study is robust construction of a generalized knowledge model, at the very least represented in the field of market research. Such a knowledge model, parametrized by the h-score, will indicate to clients the criticality of a review, which when combined with already existing metrics, we hypothesize will lead to greater actionable content.

\section{Related work and problem description}
\begin{figure*}[ht]
\centering
{
\includegraphics[width=1\linewidth]{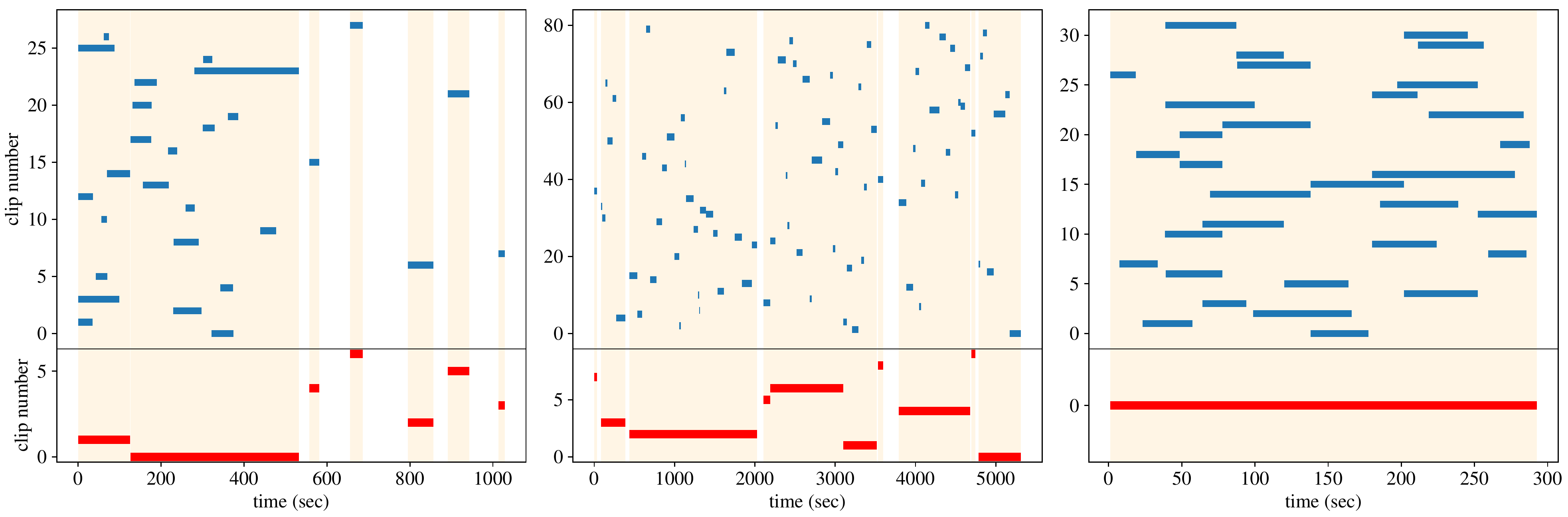}
}
\caption{Top half of plots represents individual clips (blue) as a function of time prior to superclip reduction, while bottom half displays superclips (red). Clips are separated on the y-axis, each integer representing a separate clip, for visualisation purposes.}
\label{fig2}
\end{figure*}

The problem we are studying is similar in nature to text summarization, key-phrase extraction, and topic modeling but it contains some fundamental differences we wish to clarify now.
Whatever the method employed, the ultimate goal of text summarization is to distill a large body of text into an abridged version - an abstract. Building on this, key-phrase extraction falls into a specific category of text summarization: extracting important topical words or phases from a document. Topic modeling is a somewhat different technique, requiring a large collection of documents for it to be efficient. From this large corpora latent features, or \textit{topics}, are identified and generally used to classify texts. 

The problem we are tackling, highlight extraction, cannot be addressed by text summarization methods. The solution to the problem we are addressing is neither to provide the user with an condensed summary of the document, nor to generate any classification metrics. The goal is to extract the most \textit{interesting} fragment of text, as judged by a wide variety of subjective human opinions.  The closest analog in text analysis is extraction of quotes from a novel. The quotes do not typically provide the reader with a summary, nor are they used to classify the books in question. Instead, they give the reader a vignette of the most important fragments of the text. These important moments are what we refer to as \textit{highlights}.

The concept of a highlight is most prevalent, and likely originated, in the broadcast sports industry, where it's possible to generate a precise definition of highlight. As such, there are multiple successful supervised approaches employing heuristic rules in the field \cite{Otsuka_2005,Rui_2000,Tong_2005,Zhang_2006}. Noting the limited scope of rule-based approaches, recent research has attempted to create more general models with considerable success. Yang et. al. created a robust auto-encoder model for extracting moments of interest from YouTube video data \cite{YangWLWGG15}. Bettadapura et. al., on the other hand developed techniques to evaluate and score the aesthetic composition of vacation home-video data, on top of additional contextual information such as GPS data \cite{vacation_highlight22}. Highlight extraction has also been attempted in music by Ha et. al. who used a CNN approach, incorporating an attention based component, to analyze music data and generate automatic free previews of songs. \cite{music_highlight}. 

To the best knowledge of the authors there is little to no work done on highlight extraction from text data. On the other hand, there are web based platforms which proclaim similar analysis but they rely heavily on a heuristic approaches. This consists of layering various time-series quantities on top of content - sentiment scores, keyword and entity extraction, etc -  and manually generating a set of rules based on the type of content. The technique is more generally referred to as template instantiation and is studied extensively in \cite{frump82,autosum}. These approaches are very effective for certain problems\cite{slaton1996,paice1990}, however the techniques do not posses domain invariance and the rules must be rebuilt for each new problem. In contrast, in our work, we do not manually generate any hard-coded logic or lexicons and seek to crate a robust, automated, highlight extractor.

\subsection{Classifiers as representations of knowledge models}

We adopt the formulaic approach to dynamic knowledge creation proposed by \cite{knowledge_models}. Nonaka defines two types of knowledge:
\begin{quote}
    Explicit knowledge can be expressed in formal and systematic language and shared in the form of data, scientific formulae, specification, manuals and such like. In can be processed, transmitted and stored relatively easily. In contrast, tacit knowledge is highly personal and hard to formalize. Subjective insights, intuitions and hunches fall into this category of knowledge. Tacit knowledge is deeply rooted in action, procedures, routines, commitment, ideals values and emotions. [...] It is difficult to communicate tacit knowledge to others, since it is an analogue process that requires a kind of "simultaneous processing".
\end{quote}

He goes on to argue that knowledge models are what enable true innovation within organizations. They separate subject matter experts from the newly initiated and innovative from monolithic. In a broad sense, Nonaka proposes that knowledge models are at maximum value when tacit knowledge, is blended with record based explicit knowledge. The blending serves as the basis for confidence in decision making, enabling risk taking and innovation on an individual level as well as organizational. 

The main challenge is that is they are largely based on environmental and social observation - they are highly personal and hard to crystallize. There are numerous perspectives within organizational theory on how these models are created, broken down, managed and externalized. There is unanimous agreement on their high value but few organizations are able to leverage them at scale. Focusing on the tacit knowledge within the market research industry we set out to architect and train machine learning algorithms to embody the market research industries' tacit knowledge. Using these models we then address one of the biggest and most universal problems within the market research industry: cost and time associated with video analysis.

Within this context we can understand why highlights are difficult to define - precisely because the "insights" of a sentence is in fact tacit knowledge. The analog exists in literature: why is one sentence worthy of being quoted and another is not? In literature highlights, such a model might place emphasis the profundity of the statement, the ability to stir up some feeling in the reader, or the usage of expressions in a novel and curious way. Such factors, and many more, might constitute a literature tacit knowledge model. 

To understand how a highlight classifier can be viewed as a knowledge model one must look to the data and what it represents. The majority of the semantic content consists of users expressing opinions. Market researchers approach this semantic content with questions like; what is critical in the sense of possessing high predictive power? Which of them lead to the most actionable insight for the business? The process researchers undertake when analyzing content is to sift through videos and chunk out regions of content which they they believe will answer these questions. This process is driven by their underlying tacit knowledge models, constructed over the course of many years of experience. In this light, building a classier to mimic the highlight extraction process can then be seen as the aggregation and implementation of \textbf{all} the tacit models of the market researchers. 

Although we expect that a fraction of the data will be noise, the fundamental assumptions of this study are: i) the clip-selection process is predicated on the existence of one or more underlying knowledge models and that ii) the clips researchers actively and intelligently chose contain intrinsic value. Armed with these assumptions we formulate our hypothesis:
\begin{quote}
\textbf{Hypothesis: There is a semantic difference between a highlight and a non-highlight.}
\end{quote}
Our goal is to argue convincingly that not only is this true, but that this difference can also be modeled - and modeled with high accuracy. We begin with a discussion of the proposed methodology. 

\section{Our Approach}

In this paper we study several binary highlight classification models for text evaluation:
\begin{equation}
\Lambda_k\left(\{ w_1, w_2, w_3 \cdots w_j\}\right) = y_i
\end{equation}
Here $\Lambda_k(X_i)$ is a non-injective, non-surjective function which takes as input an ordered set of words and returns a binary score, $y_i\in\{0,1\}$. The index $k$ represents the class of model to which $\Lambda$ belongs and $\Lambda$ can be understood to represent all of the respective models' parameters and transformations \cite{wembed1} . In this study we consider Gradient Boosting Models ($k=$GBM) with an implementation of LightGBM as well Recurrent Neural Networks ($k=$RNN) with LSTM architecture. In the case of RNN models the index $j$ is the maximum sequence length - 400 words - past which we truncate. The goal of this work is to optimize a highlight regressor model, s.t. $y_i \in [-1,1]$ where $y_i$ is the continuous discriminator variable corresponding to a particular highlight classification model (GBM or RNN). Thus $y_i$ has the interpretation of a similarity metric that scores the "highlight-likeness" of a given clip. The first main result of this study is a model, $\Lambda(X_i)$ that outputs a score for how likely a given clip is to be a highlight, which we refer to as the highlight score, or \textit{h-score} for short.%Q1 heavy rewriting

Special attention must be paid to the dependence of model performance on the data sampling method. This is due to the inherent ambiguity for how a large transcript should be sliced into short segments corresponding to coherent, standalone fragments. Our approach is to design four rudimentary algorithms which sample with replacement fragments from a large document. The min and max length of these fragments are chosen such that 99 \% of user generated clips fall in this range. The fragments are then evaluated with the best performing classifiers and pieced back together to form the original document. We design a schema for rating the sampling algorithms by comparing the highest scoring fragments with human opinions.%Q1 heavy rewriting

However, the first significant challenge we must face is to construct a dataset suitable for a machine learning exercise. We turn to this task now. 

\begin{table*}[t]
\centering
\begin{tabular}{lccccc}
{} &     transcripts &      clips\_orig &     clips\_clean &          nclips &   nclips\_renorm \\
\hline
count  &         138,799 &         586,062 &         328,070 &         121,210 &       543,130 \\
mean   &        680 (53) &          89 (6) &          88 (6) &        311 (27) &        69 (6) \\
median &        347 (22) &          67 (4) &          67 (4) &          80 (6) &        48 (4) \\
mode   &        124 (10) &          42 (2) &          41 (2) &           0 (1) &        21 (3) \\
std    &     1,437 (143) &        106 (10) &        108 (10) &     1,064 (108) &        79 (6) \\
min    &           1 (0) &           0 (0) &           1 (0) &           0 (0) &         1 (1) \\
25\%    &        175 (12) &          39 (2) &          39 (2) &          18 (2) &        24 (3) \\
75\%    &        653 (42) &         111 (7) &         110 (7) &        240 (17) &        87 (7) \\
max    &  63,573 (7,373) &  11,391 (1,247) &  11,391 (1,156) &  62,627 (7,260) &   3,267 (280) \\
\hline
\end{tabular}
\caption{Word counts (sentence counts) along with statistical quantites of interest for the steps of the raw data treatment.}
\label{tab:1}
\end{table*}

\section{Dataset engineering}
\label{de}
Our dataset contains over 130,000 videos and accompanying transcripts, plus metadata. Word count and sentence distributions are shown in Figure. \ref{fig3}  and accompanying statistics in Table \ref{tab:1}.  Over $99\%$ of the data is human transcribed and is formatted according to multiple schemes, Fig. \ref{fig1}, depending on the transcription service that was used. Many of the long-form transcripts contain annotations (time stamps, speaker tags, annotations). Accompanying each transcript is a complete list of user-generated clips with accompanying metadata. The parameters most relevant to our analysis are the clip start and end times. %Q1 removed unnecessary information about the annotations

\subsection{Pre-processing}%Q1 moved from section "Model Design"

We investigate the performance of the models as functions of two parsing techniques: inclusion of punctuation and stemming. We theorize that in addition to word choice, punctuation has the ability to convey meaningful information in regards to the text content. We base this assumption on the usage of punctuation in sentiment analysis, where it can be shown that happy vs furious customers will use significantly different ratios. Stemming, on the other hand, is a normalization technique that removes variation from text by trimming morphological variation. If model performance doesn't suffer greatly due to stemming, it will allow us to build much more efficient estimators. The data also undergoes standard cleaning procedures such as: removal of duplicates, NaNs, and corrupt entries, removal of stop words, stemming, lowercasing, and tokenization depending on the model.%Q1 sentence moved from first paragraph of section "Dataset Engineering" %Q1 removed sentence "This should help define the boundary between highlights/non-highlights." %Q1 removed "We expect the inclusion of punctuation to make a significant difference in RNN models specifically. "

\subsection{Clip Segmentation}%Q1 added subsection "Clip Segmentation" to ingest the following two related subsubsections

In order to collect evidence for our hypothesis we must create a training/test set which exhibits clear separation of clips and non-clips. A high degree of redundancy in the dataset is observed since multiple clips were often created from the same portion of video, see Figure \ref{fig2}. This is a variant of the known shortest common sequence problem, and we solve it by writing a greedy algorithm which transverses the list of clips and generates superclips. %Q1 changed "prove" to "collect evidence"

\subsubsection{Shortest superclip reduction}
Consider an analog with strings. From the set
\begin{equation}
    S = \{abc, cde, fab, hij, ab, a\} 
\end{equation}
we construct the superstring set 
\begin{equation}
    \tilde{S} = \{fabcde, hij\}
\end{equation}
such that every element of $\tilde{S}$ is composed of overlapping or individual elements of $S$. Results of the algorithm can be visualized in Figure \ref{fig2} on three random transcripts. Applying this process to the entire data-set leads to a drastic reduction in clip number. Statistical descriptors in of clips before superclip reduction are in Table \ref{tab:1} column \textbf{clips\_orig}, while  column \textbf{clips\_clean} is post-treatment. 

While there are no major changes in the averages and quartiles the number of clips is reduced to almost half. There are multiple reasons for this high degree of redundancy: many researchers analyzing the same video, user errors, project redundancy. Whether or not superclip reduction is necessary in constructing the data-set is still up for debate. It can be argued that superclips resulting from multiple overlaps should be weighted stronger than one which doesn't, since they are indicative of important semantic content. We do not take this approach and weigh all superclips equally. Construction of superclips is necessary to achieve clear non-overlapping boundary between clips/non-clips by string excision. 

\begin{figure}[ht]
\centering
{
\includegraphics[width=1\linewidth]{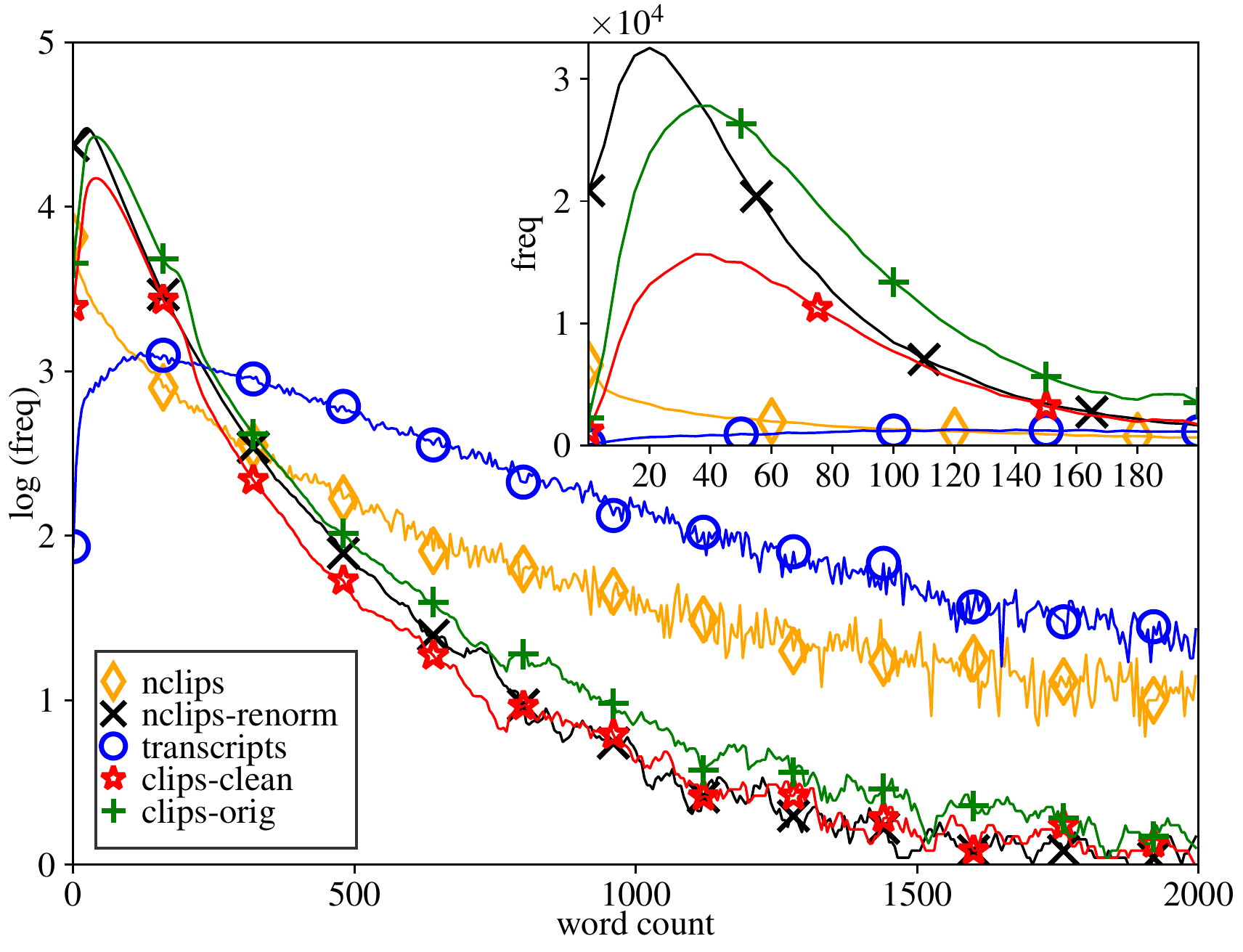}
}
\caption{Frequency as a function of word count, on a logarithmic scale, of dataset at various points in the clening process. }
\label{fig3}
\end{figure}

\subsubsection{Redistribution}
The superclips now excised from their corresponding transcripts leave behind what we refer to as the non-clips (\textbf{nclips} Figure \ref{tab:1}). There is a significant covariate shift between the two classes apparent in the distributions of \textbf{clips-clean} and \textbf{nclips}. We use a modified Kolmogorov Smirnov metric to measure this shift
\begin{equation}
    D(\bar{X}_{c}, \bar{X}_{nc}) = \int_{0}^{\infty} \left|p_{X_i}^{nc}(x) - p_{X_i}^{c}(x)\right| dx
\end{equation}
where $p^{c}$ ($p^{nc}$) are the probability densities of the clips (non-clips) plotted in Figure \ref{fig3}. While this does not pose as serious a problem as the covariate shift occurring between train and test set data \cite{covshift}, we are inadvertently introducing into into the model $\Lambda_k(X_i)$ a strong highlight bias towards short input sequences. We wish to eliminate sequence-length as a model parameter all-together, since there is no \textit{a apriori} information to indicate the contrary. 

To to this we must minimize $D(\bar{X}_{c}, \bar{X}_{nc})$ which is achieved by a string redistribution schema outlined in Algorithm \ref{alg1}. It consists of an approximate numerical minimization based on randomly sampling the problematic data with a weighted probability according to the desirable distribution and re-constructing the set. In doing this, any single transcript may become fragmented into multiple non-clips - 5 on average - of varying word counts which increases our non-clip counts to almost 550k. The sampling is always done sequentially as to not lose the semantic content of the text. The result is more than satisfactory since for the purposes of RNN modeling, since the smallest unit of text is a sentence. 
\begin{algorithm}[ht]
\SetAlgoLined
\KwData{Non-clip data $\textbf{X}^{(nc)} = \{X_i^{(nc)} |X_i \sim p_{nc}(X_i^{nc}) \}$}
\KwResult{$\tilde{\textbf{X}}^{(nc)} = \{\tilde{X}_i^{(nc)} |X_i \sim p_{c}(X_i^{c}) + \delta D, \textbf{min}(\delta D)\}$ }
Define empty set $\tilde{\textbf{X}}^{(nc)}$\;
 \For{$X_i \in \{X_i^{(nc)}\}$}{
  Define empty set K\;
  \While{$\sum(K) <$ WordCount($X_i$)}{
   add to K int sampled form $p_{c}(X_i^{c})$\;
   }
    $ind_f = [CumSum(K)]$\;
    $ind_s = [0,ind_f]$\;
   {
  \For{$(ind_s,ind_f)$ in K}{
  add to $\tilde{\textbf{X}}^{(nc)}$ sequentially $ X_i[ind_s:ind_f]$
  }
  }
 }
\caption{Redistribution of non-clips }
\label{alg1}
\end{algorithm}
Algorithm \ref{alg1}  reduces D from $13.25 \rightarrow 1.33$ ($\sim $ 90\% reduction) computed in terms of probability density and not $\log(freq)$ which are the units of Figure \ref{fig3}. 

To determine if we have removed sequence-length drift between the clips and non-clips we construct a random forest classifier and train it to distinguish between the two classes based on sequence-length alone. As a measure we chose ROC-AUC since it is immune to data imbalance. It scores 0.683 when considering the superclips (\textbf{clips\_clean}) and the original non-clips (\textbf{nclips}) in Table \ref{tab:1}, indicating a noticeable drift - typically 0.8 is considered a "strong" shift. Post re-distribution (\textbf{nclips\_renorm}) the same procedure yields 0.503, indicating that we have successfully removed all dependence on sequence length and covariate shift from the dataset. Satisfied with this result, we combine \textbf{nclips\_renorm} with \textbf{clips\_clean} to obtain our finalized dataset.

%Q1 eliminated section "Model Design" and moved the first paragraph of the previously existing sentence to the section "Experiments"

%Q1 eliminated section "Methodology"

\section{Experimental Design}%Q1 renamed section "Experiments" to "Estimator Models" and demoted to subsection

The novelty of the problem and the dataset poses a significant challenge in model selection. We chose to explore two classes of models known for their good performance on a wide range of NLP tasks. In the GBM models, the pre-processing steps include tokenization, stemming, stop-word removal, and embedding as indicated while in the RNNs we do not tokenize, stem, or remove stop-words. %Q1 added "performance (on a wide range of) NLP tasks"

\subsection{GBM models}
Non-linear tree models use the word count statistics as features. In this study we explored several common choices for word-count features, including stemmed-Bag-of-Words (SBOW), Bag-of-N-grams (BONG), and word embeddings. A brief pedagogical description of the features considered here is given in the Appendix. For each combination of pre-processing steps and feature extraction, we employ a 7-fold cross validation scheme. All models are combined to generate an ensemble model to avoid over-fitting. The ensemble model is used to determine the best threshold cutoff between the two classes and is then evaluated on the test set. In all cases we train until no improvement in AUC occurs over 40 iterations.%Q1 included definitions of the acronyms like BOW and BONG as well as references to the appnendix

\subsubsection{BONG and TFIDF + BONG}
The counts of the top 120,000 most frequent words (post stop word removal) from the training set are used as features. In TFIDF schema the counts are weighted by an appropriate inverse document frequency term. N-gram models up to a maximum of tri-grams are constructed - consequently we allow the feature space to grow in dimension to $\sim$ 1M n-grams. 

\subsubsection{SBOW and TFIDF + SBOW}
The most frequent 120,000 stemmed words are extracted (only uni-grams) and both the CV or TFIDF schemes are used to generate a vector space model. Stemming is done with PorterStemmer from NLTK. \cite{stemmer}

\subsubsection{Bag of TFIDF Embeddings}
Features are constructed via an averaging of embedding vectors according to Eq. \ref{eq:bom1} with $c_i = 1$ or via the IDF schema detailed in Eq. \ref{eq:bom2}. The dimension of the vector space is thus determined by the dimension of the embedding matrix - GloVe: 50, BlazingText: 200.

\subsection{LSTM models}
We consider three LSTM models: single layer uni-directional, single layer bi-directional, and single layer with an attention component. A brief descriptoin of LSTM models is given in the appendix. One round of hyperparameter tuning was done to determine the best regularization methods/coefficients and address the inherent noise in the data. Training of the LSTMs was stopped when two consecutive epochs did not yield an improvement in ROC AUC. As a result, models may have been trained for vastly different times, and epoch numbers. For training, we used a constant mini-batch size of 128 and ADAM gradient descent optimizer with a learning rate of 0.001.%Q1 added reference to appendix

\subsubsection{LSTM - Unidirectional} 
An input sequence maximum of 400 words is imposed. Anything shorter is padded with zeros. GloVe and BlazingText embedding matrices are used to project words into a non-orthogonal vector space. Via time evolution, each later hidden state is computed by a combination of it's corresponding word and the previous hidden state. Only the last hidden state in the sequence is then used for making predictions. \cite{deeplearning}

\subsubsection{BD-LSTM} 
A second LSTM layer with time going in reverse is added to the model. The forward and backward propagating layers do not interact with one another, so the model is trained just as before, except it is now doubled in complexity. \cite{bdlstm}

\subsubsection{A-LSTM} 
Instead of using only the last hidden unit to make inferences at prediction time, attention mechanisms allow the model to look at all the hidden units in a sequence. Simultaneously we measure the amount of emphasis the model places on each vector - the attention - which roughly translates to word attention. \cite{attention2,attention3}

\subsection{Sampling methods}%Q1 added subsection "Sampling Methods" and imported the theory-description from the section "Methodology"

In this section we discuss the algorithmic sampling methods for large transcripts. We begin by expressing a document, D as the exhaustive set of fragments, where a fragment is composed of one or more sequentially sampled sentences.
\begin{equation}
D = \{(f_i,f_{i+j}) | 0 < i < N, n_i<j \leq n_f, i+j \leq N \}
\label{eq:frag_set}
\end{equation}
here $n_f,n_i$ are the minimum and maximum fragment/clip lengths and $f_{i,j}$ are start/end sentence indexes of the respective fragment. For example, the fragment $(f_3,f_5)$ is composed of sentences 3 through 5 in order. Statistical inference from Table \ref{tab:1} indicates we should pick a range somewhere around the median/mode number of sentences $n_{i,f} \approx (3,10)$. The size of D increases linearly with transcript length. 
\begin{equation}
    dim(D) \approx N (n_f - n_i) 
\end{equation}
For reference a 1 hour long video roughly corresponds to 1000 sentences. However, we investigate a much larger range and set $n_f = 20$. In practice, evaluating the entire set in Eq. \ref{eq:frag_set} is embarrassingly redundant since there will be roughly a factor of $n_f - n_i$ overlap between the fragments. We thus propose four different sampling algorithms, which aim to reduce this computational complexity as much as possible while intelligently eliminating all existing overlap.

\subsubsection{Sampling algorithm I. Sequential sentence based method (eq11) } This is the simplest approach boiling down to evaluating each sentence as a standalone fragment. This corresponds to $n_i, n_f = 0,1$ and Eq. \ref{eq:frag_set}
becomes
\begin{equation}
D = \{(n_i,n_{i+1}) | i+1 \leq N \}
\label{eq:frag_set1}
\end{equation}
This approch loses a significant amount of content, since we are imposing on each sentence a forgetfulness property i.e. not dependent on any previous or following state. With this algorithm, the smallest and largest highlight clip we are able to generate is one sentence. On the other hand, the redundancy of Eq. \ref{eq:frag_set} is removed. 

\subsubsection{Sampling algorithm II. H-score non-overlap (hscore) }
This approach incorporates overlapping fragments. First generate an empty set Q and
\begin{itemize}
    \item Score all the fragments in Eq. \ref{eq:frag_set} 
    \item Sort fragments by h-score
    \item Move down the sorted list (in descending order) and store highest scoring entry in Q only if it doesn't overlap with any entries already in Q.
\end{itemize}
This guarantees no overlap between the elements of this new set, however it doesn't guarantee full coverage of original transcript. 

\subsubsection{Sampling algorithm III. Weighted H-score non-overlap (hscorelength)}
It is possible longer highlight clips are desirable in certain circumstances. We introduce an additional parameter, a weight to the h-score of each fragment which will prioritize longer fragments over shorter ones. We chose the modified h-score to be
\begin{equation}
    \bar{h_{s}} = h_{s} * (n_f - n_i)
\end{equation}
The same procedure as Sampling Algorithm II. is followed with $\bar{h_{s}}$ instead of h-score.

\subsubsection{Sampling algorithm IV. Positive h-score summation (possum)}
A final, more robust method is inspired by \cite{videosum}. Fragments are scored and all the negative scoring ones are dropped. To generate predictions we employ an h-score averaging scheme according to 
\begin{equation}
    s_i = \sum_i \frac{s\{f_i\}}{n} 
\end{equation}
where $s\{f_i\}$ is the set of scores of every fragment which contains the $i^{th}$ sentence. The score of each sentence $s_i$ is a a mean field h-score. 

\subsection{Performance evaluation}%Q1 renamed subsection "Part I. Evaluating estimator performance" to section "Performance Evaluation"

\begin{figure}[ht]
\centering
{
\includegraphics[width=1\linewidth]{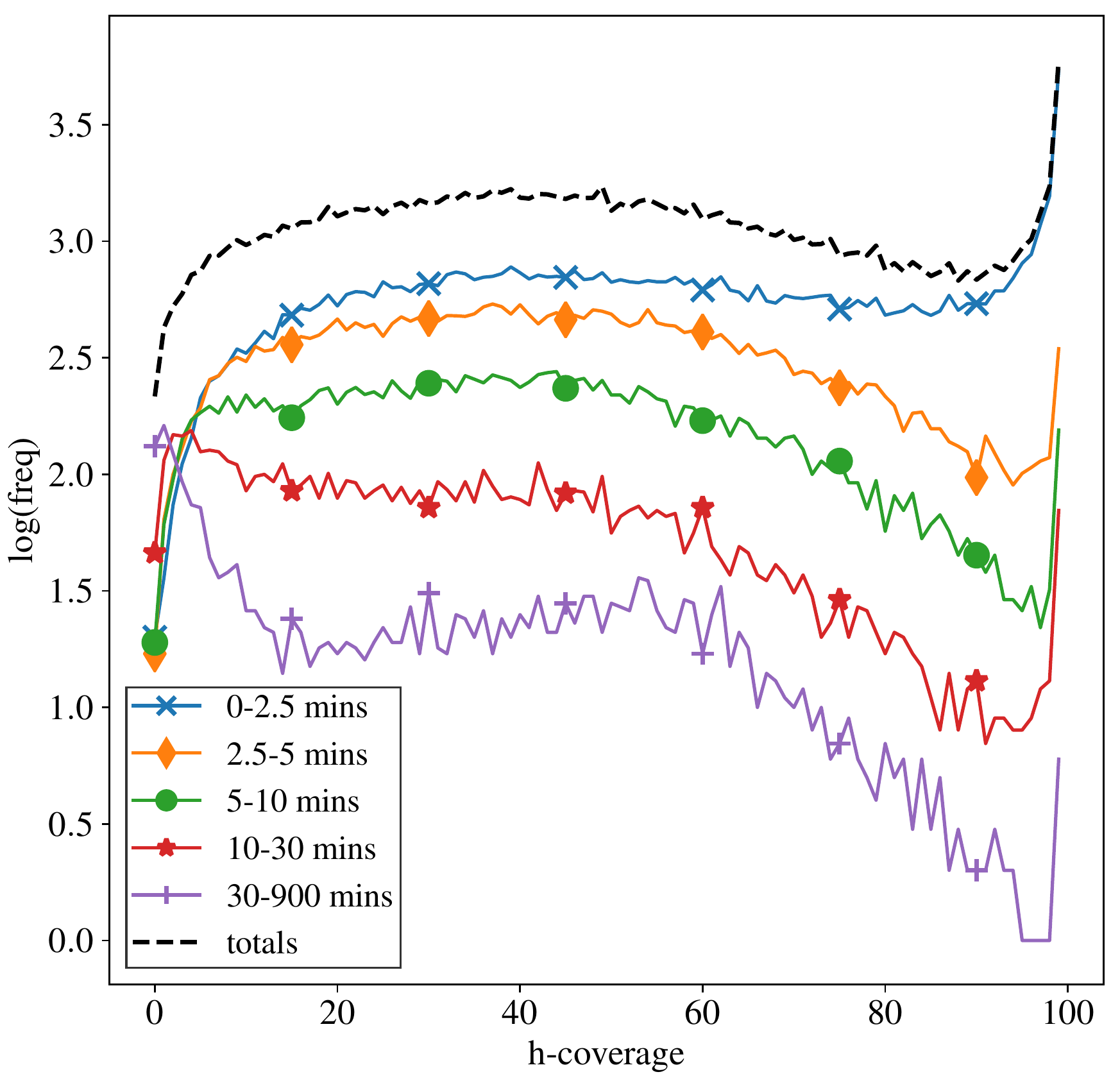}
}
\caption{Logarithmic distribution of transcripts counts plotted against h-coverage. Data-set is split by length of the transcripts in minutes to illustrate different phenomena occurring for shorter and longer videos. The dashed line represents the summation of all others. Approximately 3000 transcripts have an h-coverage of 100\%, most of them being on the shorter end.}
\label{fig4}
\end{figure}

A train/dev/test split was generated of $\sim$ 800k/40k/40k clips. The split was performed to ensure sampling obeys a constant distribution with respect to \textit{h-coverage} - the percentage of the document which users turn into clips. The peak near 100\% in Figure \ref{fig4} indicates that shorter videos are much more likely to become highlights in their entirety. The converse is true for long videos, where h-coverage is the lowest, and peaks close to zero. 

\begin{figure}[ht]
\centering
{
\includegraphics[width=1\linewidth]{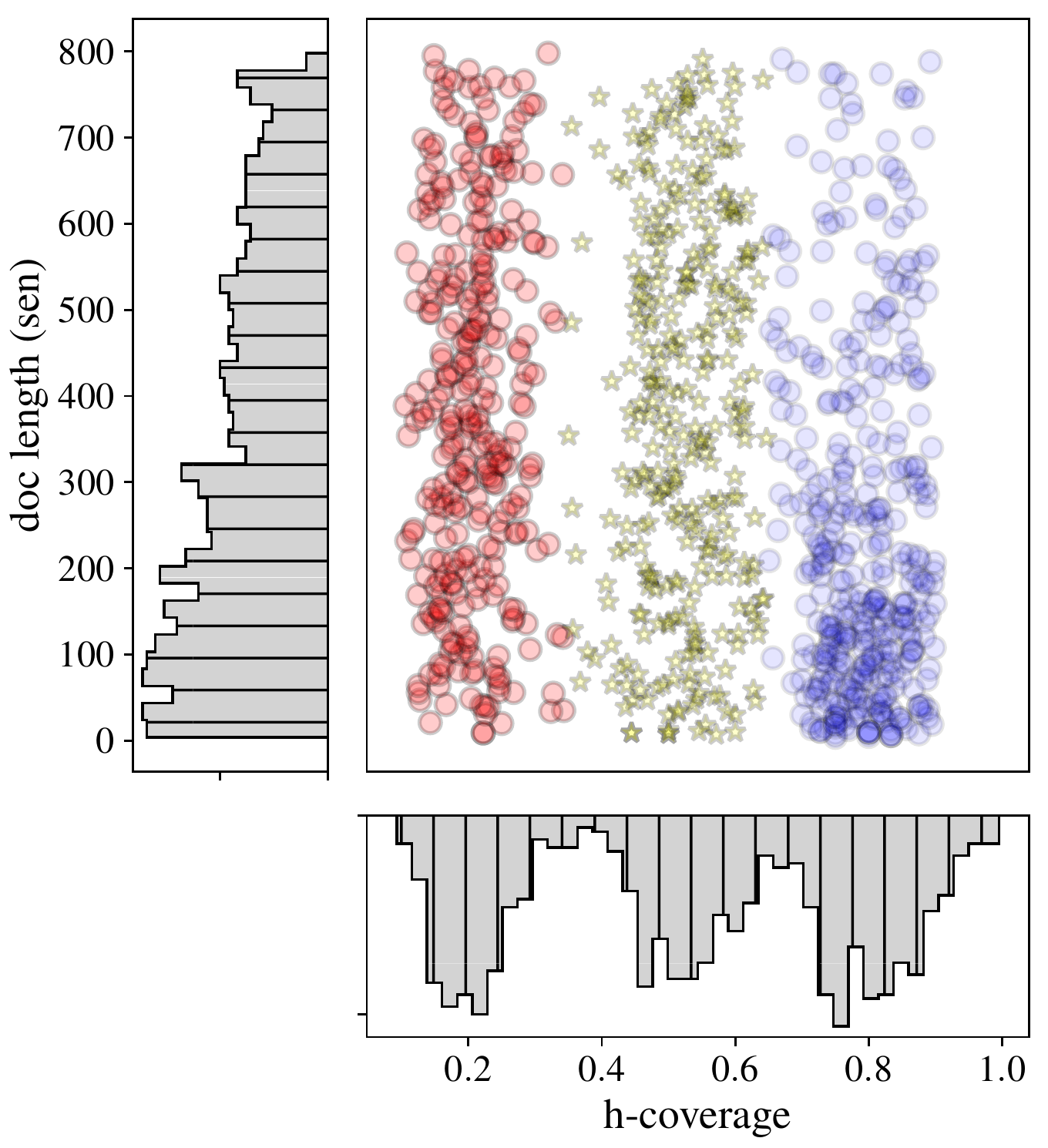}
}
\caption{Description of test dataset used for evaluation of the sampling algorithms. Each point represents a transcript. Vertical and horizontal insets are the distributions with respect to sentence counts and h-coverage. Each h-coverage cluster obeys a normal distribution with sample sizes $N_{1,2,3} \approx 375$ (h-cov $\approx 0.2)$,  305 (h-cov $\approx 0.5)$, 300(h-cov $\approx 0.8)$. Document length distribution is roughly uniform with a slight skew towards shorter videos.} 
\label{fig7}
\end{figure}

\subsubsection{Evaluating sampling algorithms}
The four sampling algorithms described in the previous section are coupled with the best LSTM (\textit{lstm\_fpa}) and GBM (\textit{tfidf\_stem}) estimators. Their performance is then evaluated on $\approx$ 1000 transcripts taken from the test set (Figure \ref{fig7}). We have chosen to investigate closely the relationship of model performance with that of document length, and h-coverage. Recall that \textit{h-coverage} refers to the percentage of highlight fragments among the total fragments of a transcript. The dataset is constructed as follows:
\begin{itemize}
    \item We sampled our test set to generate a uniform distribution with respect to document length. 
    \item Three h-coverage values were randomly sampled to form normal distributions centered at (0.2, 0.5, 0.8) with a variance of 0.1. 
\end{itemize}
We construct a truth set by segmenting every document into sentences and ii) assigning each sentence either a 0 or a 1 based on whether that sentence belongs to a user-generated highlight clip. Because a significant portion of clips begin mid-sentence, a sentence will be considered a highlight if more than 75\% of it overlaps with one existing in a clip. The total length of this test dataset is 342,000 sentences, with an average h-coverage of 0.52.

For evaluation, we use the same metric as in \cite{karen1}. Each transcript is evaluated with all sampling methods. We sort the predictions in descending order of h-score, corresponding to the likelihood of that particular sentence of being a highlight. We then compute the recall, precision, and the ROC ACU score of the top R fragments with respect to the truth set. Finally, this number is averaged across all transcripts to obtain the mean average precision, recall and ROC AUC designated by mAP, mAR and mAROC.

\begin{table}[t]
\centering
\begin{tabular}{lrrrrrrrrrr}
{} &  roc\_auc &      prec &       rec  \\
\hline
\hline
cv11       &      \textbf{0.844477} &    0.655824 &  0.774548  \\
cv12       &      0.844214 &    \textbf{0.655962} &  0.773516 \\
cv13       &      0.843592 &    0.652568 &  0.772485  \\
cv14       &      0.843555 &    0.649319 &  0.783651  \\
cv22       &      0.781970 &   0.602158 &  0.697536  \\
cv33       &      0.633714 &   0.548535 &  0.398835  \\
tfidf11    &      0.840925 &  0.646296 &  0.786867 \\
tfidf12    &      0.842626 &   0.639390 &  \textbf{0.793786} \\
tfidf13    &      0.842457 &  0.645633 &  0.785472 \\
tfidf14    &      0.841936 &   0.644063 &  0.785047 \\
tfidf22    &      0.787775 &   0.593215 &  0.745964  \\
tfidf33    &      0.638178 &   0.545203 &  0.430392  \\
\hline
mFVem      &      \textbf{0.834259} &  \textbf{0.630410} &  0.792086 \\
mGlove     &      0.767736 &   0.545318 &  \textbf{0.794150} \\
tfidfFVem  &      0.825609 &   0.624443 &  0.773941  \\
tfidfGlove &      0.753868 &   0.552547 &  0.728122  \\
\hline
cv\_stem    &      0.867260 &    \textbf{\underline{0.671149}} &  0.817818  \\
tfidf\_stem &      \textbf{\underline{0.868849}} &  0.667766 &  \textbf{\underline{0.823340}} \\
\hline
\hline
\end{tabular}
\caption{Experimental results of Gradient Boosting Highlight Classifiers (LightGBM). All values shown are for 7-fold ensemble models. Classes of models are separated by horizontal lines. Best scores form each section (overall) are in bold (underlined). Feature notation class 1: (cv/tfidf)(ab) - count vectorization/tfidf weighting with a/b: minimum/maximum n-grams. Class 2: (m/tfidf)(embedding) - mean/tfidf weighting of word embedding vectors as described in Eq. \ref{eq:bom2} with GloVe/BlazingText embeddings. Class 3: (cm/tfidf)(embedding) - count vectorization/tfidf weighting of stemmed unigrams.}
\label{tab:2}
\end{table}

\begin{table}[t]
\centering
\begin{tabular}{lrrrrrrrrrr}
{} &  roc\_auc &   prec &       rec  \\
\hline
\hline
lstm\_f    &  0.880486 &  0.743226 &  0.752029 \\
lstm\_fa   &  0.891008 &  \textbf{0.782102} &  \textbf{0.754341}  \\
lstm\_fab  &  \textbf{0.898833} &   0.772039 &  0.736791 \\
lstm\_fb   &  0.876495 &   0.781038 &  0.744875 \\
\hline
lstm\_g    &  0.869853 &    0.726934 &  0.645714 \\
lstm\_ga   &  \textbf{0.880833} &   0.725303 &  0.711028  \\
lstm\_gab  &  0.880539 &  \textbf{0.738461} &  \textbf{0.744230} \\
lstm\_gb   &  0.867619 &   0.727030 &  0.693155  \\
\hline
lstm\_fp   &  0.932866 &  0.836888 &  0.799052  \\
lstm\_fpa  &  \textbf{\underline{0.940302}} &    0.840100 &  0.806834  \\
lstm\_fpab &  0.938947 &   \textbf{\underline{0.842375}} &  0.803886  \\
lstm\_fpb  &  0.931851 &   0.817474 &  \textbf{\underline{0.818245}}  \\
\hline
lstm\_gp   &  0.920107 &  0.817944 &  0.727723  \\
lstm\_gpa  &  \textbf{0.935798} &  \textbf{0.824796} &  0.801406  \\
lstm\_gpab &  0.935498 &    0.816673 &  \textbf{0.804897}  \\
lstm\_gpb  &  0.908086 &    0.663225 &  0.545645  \\
\hline
\hline
\end{tabular}
\caption{Experimental results of LSTM Highlight Classifiers (TensorFlow). All values shown are for 7-fold ensemble models. Classes of models are separated by horizontal lines. Best scores form each section (overall) are in bold (underlined). Baseline model is a 128 hidden unit, single layer LSTM, with 64 batch size. \textbf{g}/\textbf{f} designates usage of GloVe/BlazingText embeddings, \textbf{a} - attention mechanism, \textbf{b} - bi-directionality, \textbf{p} - use of punctuated dataset for training. }
\label{tab:3}
\end{table}

\begin{figure*}[ht]
\centering
{
\includegraphics[width=1\linewidth]{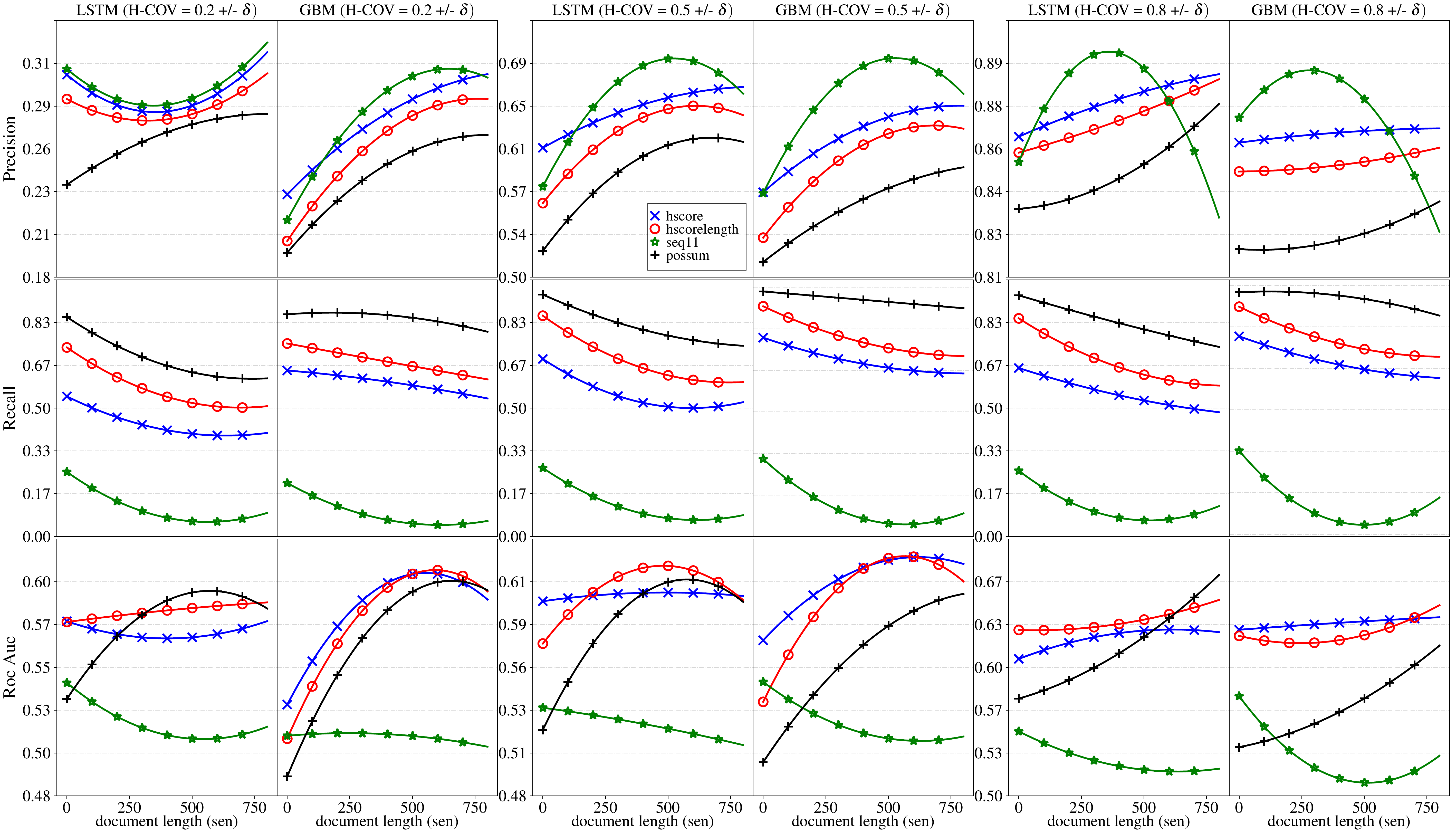}
}
\caption{Experimental results of the four sampling algorithms evaluated with the two best performing GBM and LSTM models from Tables \ref{tab:2} and \ref{tab:3}. Precision, recall and ROC AUC scores are illustrated for h-coverages of 0.2, 0.5, 0.8 and as continuous functions of document lengths. Test dataset described in Figure. \ref{fig7}. }
\label{fig6s}
\end{figure*}

\section{Results and Discussion}
We compare the LightGBM and LSTM based models on multiple performance metrics to point out different strengths and weaknesses. We do not focus on the hyper-parameter tuning procedures and discuss only the best models from each category. For the sampling study, the best two models are used to evaluate the fragmenting algorithms. All experiments are implemented in Python on a workstation with 4x Nvidia 2080-TI graphics cards, and 12-core 3.1GHz CPU. Cumulative model training, including hyper-parameter tuning, amounts to approximately 1 month of total training time. When it was possible the training was done on the GPUs to take advantage of CUDA core parallelization. 

%Q1 removed section "Comparison of LSTM and GBM classifiers"

\subsection{Effects of features selection} 
Tables \ref{tab:2} and \ref{tab:3} summarize the performance of the standalone estimators - notation is discussed in the captions. Both tables show detailed comparisons of the various models as well as the scale invariant and imbalance-robust metric: ROC AUC.

One immediate and surprising result is the similar performance of count vectorization and TFIDF weighting. To understand the mechanism of disambiguation by the BOW model we must assume that common words are just as important, if not more than rare words. Taking the result at face value indicates that common words, words which appear in multiple documents, are the true driving force behind this effect. We can surmise they are best at distinguishing between the classes. Albeit a counter-intuitive result, one possible explanation is closely tied to human psychology. There is substantial research \cite{psycho,psycho2} that shows the use of some function words i.e. first person singular pronouns "I", change depending on someone's psychological state. For this reason it is possible that such segments of focus groups are most indicative of highlights. 

An alternate explanation is that TFIDF smoothing works against us. While the classifier is trying to draw an accurate boundary between highlight and non-highlight clusters the smoothing of the IDF term makes this boundary less fine grained. IDF is designed to spot similarity between one sample and the rest of the corpus, but we are trying to look for similarities between one sample and other clusters. Further evidence to this point is supported by the fact that increasing BONG model complexity does little to improve its performance. All other n-gram models perform just as well, if not poorer, than the uni-gram model. 

It can be seen from Table \ref{tab:2} that the bag of embedding models offer very little performance improvement, 0.793 vs. 0.784 sensitivity. When weighted against the number of features in the models - 300 for BlazingText, 50 for GloVe \cite{glove1,glove2} and 50,000 - 1M for BOW models - the performance is impressive. We no longer have the model relying on individual words, but instead on word classes (clusters of similar context). The character of each of these clusters is projected onto one of the 50 (300) latent dimensions of our vector space, which more often than not, don't correspond to a human interpretable semantic meaning.   

The stemmed uni-gram models performed significantly better than all other approaches (Table \ref{tab:2}). Because some words exist with multiple variations, depending on the grammar, it is sometimes good practice to normalize them to their absolute roots. For example, the words \textit{playing}, \textit{players}, \textit{played} all become \textit{play}. As far as the classifier is concerned its performance will be better when there is a single feature for N variations, rather than N features for each variation. Since our classifiers improve when stemming is applied, it indicates we are doing just this - reducing the number of useless features into fewer, stronger ones.

\subsection{Effects of model class}
The main difference between LSTM and GBM models is the element of time dependence within the data. LSTM's can capture time-sensitive information with their memory gates and it comes as no surprise that they greatly outperform GBM methods. 

The improvement in performance exists regardless of the LSTM variant used. Even the worst LSTM performs better than the best GBM (Table \ref{tab:3}). Among the LSTM variants themselves, we point out some obvious trends. Bidirectional LSTMs are very useful where the context of the input is needed, for example in situations where performance might be enhanced by knowledge of what comes before and after a specific point in time. We see no increase from bi-directionality at all, and in fact, in some cases we see a small decline in performance. This could occur due to the increased complexity of the network and the difficulty in reaching the local minimum. If we also account for practical aspects, the training time doubles and saturation occurs much slower.

\begin{figure}[ht]
\centering
{
\includegraphics[width=1\linewidth]{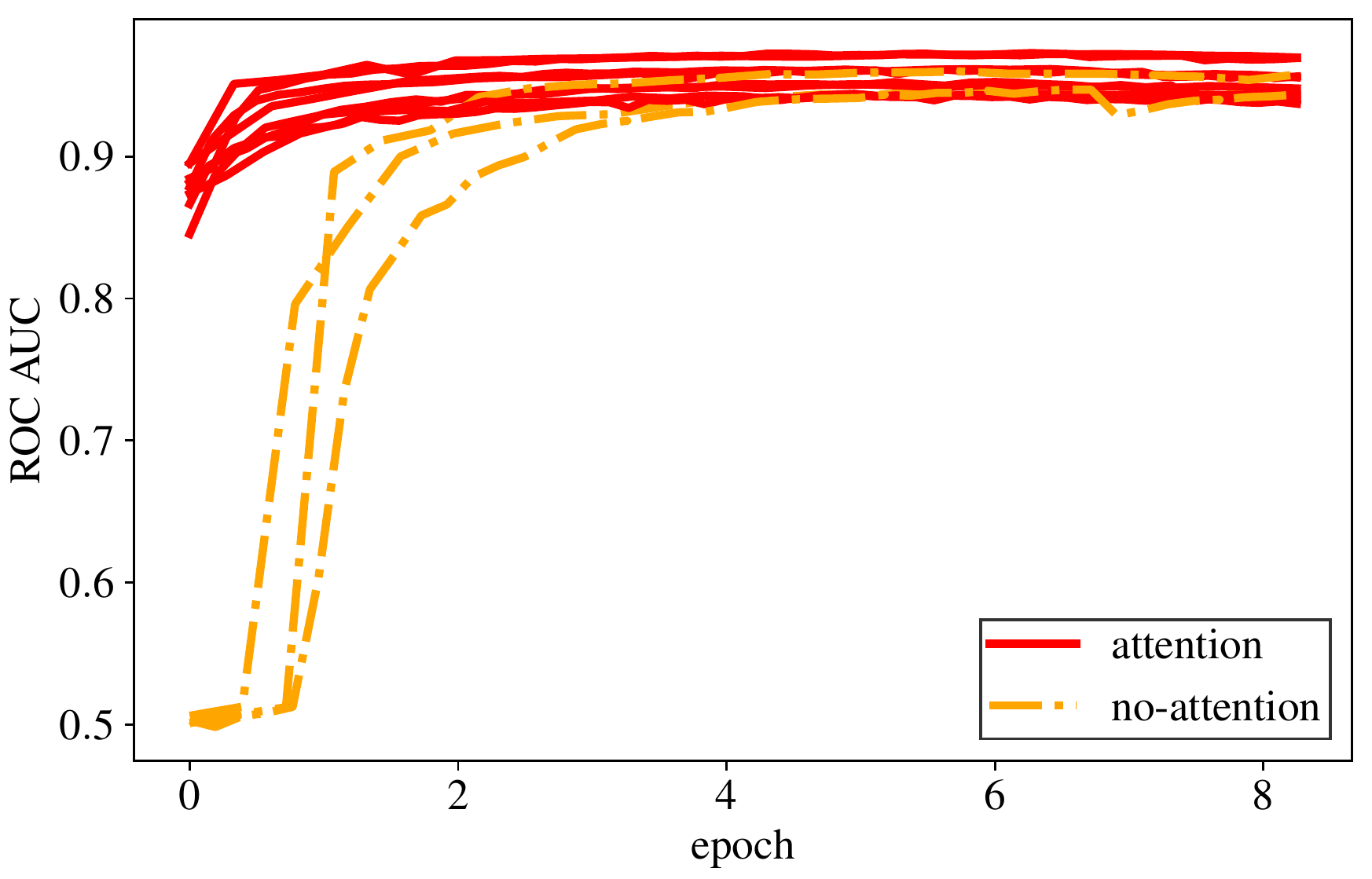}
}
\caption{ROC AUC scores (evaluated on dev set) of LSTM models as a function of training epoch. Architectures with attention components (solid red) achieve best discrimination thresholds significantly quicker than non-attention models (dashed yellow).}
\label{fig5}
\end{figure}

The addition of attention, on the other hand, improves ROC AUC scores by $\sim$ 1\% across the board. Though model complexity only increases slightly, the training time decreases dramatically. With non-attention models, diminishing returns to ROC AUC start being noticeable after 7-10 epochs, while in attention models this drops down to 2-3 epochs as seen in Figure \ref{fig5}. It is conceivable that allowing the context vector access to every hidden state generates better descent vectors which leads faster learning since crucial words (or groups of words) in the decision making process are picked up much faster.

From Table \ref{tab:3} we can see that neural network based methods that do not take advantage of personalized embeddings systematically under-perform. Training a distributed representation of words has been shown before to be very sensitive to the corpora it is trained on. \cite{embeddings2,embeddings3} Since every vector encodes semantic relationships among words, it is the context of those words which decides their representation. Training an embedding space on corpora about biology might show a high similarity score between "\textit{apple}" and "\textit{pear}" while this might not be the case if trained on a more technological corpus. In a similar manner, the custom-trained embedding space will better capture the unique terminology to the market research corpora. The performance increase is most significant in terms of precision, increasing from 0.72 to 0.78, while recall increases from 0.70 to 0.73.

\begin{table*}[t]
\centering
\begin{tabular}{|*{8}{c|}}  % repeats {c|} 18 times
\hline
& & \multicolumn{3}{|c}{LSTM Models} & \multicolumn{3}{|c|}{GBM Models} \\ \hline
& \diagbox[innerwidth=2cm]{h-cov}{doc len}  & \multicolumn{1}{|c}{$<$200 words} & \multicolumn{1}{|c}{200-500 words} & \multicolumn{1}{|c}{$>$500 words} & 
\multicolumn{1}{|c}{$<$200 words} & \multicolumn{1}{|c}{200-500 words} & \multicolumn{1}{|c|}{$>$500 words} \\ \hline 
\multirow{3}{*}{hs} & 0.2 & 0.32/0.54 &  0.28/0.45 &   0.3/0.41 &  0.25/0.64 &  0.27/0.64 &   0.3/0.59 \\
\cline{2-8}
 & 0.5 & 0.61/0.69 &  0.62/0.55 &  0.65/0.52 &  0.57/0.77 &   0.6/0.72 &  0.63/0.68 \\
\cline{2-8}
  & 0.8 & 0.86/0.66 &  0.87/0.57 &  0.86/0.55 &   0.85/0.8 &  0.86/0.71 &  0.85/0.68 \\
\hline
\multirow{3}{*}{hsl} & 0.2 & 0.3/0.73 &   0.28/0.6 &   0.3/0.53 &  0.22/0.75 &  0.25/0.72 &  0.29/0.66 \\
\cline{2-8}
 & 0.5 & 0.57/0.85 &    0.6/0.7 &  0.63/0.64 &  0.53/0.89 &  0.57/0.81 &  0.61/0.76 \\
\cline{2-8}
  & 0.8 & 0.85/0.84 &  0.86/0.72 &  0.85/0.66 &  0.84/0.91 &  0.85/0.82 &  0.83/0.77 \\
\hline
\multirow{3}{*}{seq11} & 0.2 & 0.33/0.24 &  0.28/0.12 &  0.31/0.07 &  0.22/0.19 &  0.29/0.12 &  0.31/0.06 \\
\cline{2-8}
 & 0.5 & 0.58/0.24 &  0.65/0.12 &  0.67/0.08 &  0.55/0.28 &  0.63/0.12 &  0.68/0.07 \\
\cline{2-8}
  & 0.8 & 0.85/0.24 &  0.89/0.13 &  0.87/0.09 &   0.87/0.3 &  0.88/0.13 &  0.86/0.08 \\
\hline
\multirow{3}{*}{possum} & 0.2 &0.25/0.83 &  0.27/0.73 &  0.28/0.65 &  0.21/0.89 &  0.23/0.85 &  0.27/0.86 \\
\cline{2-8}
 & 0.5 & 0.52/0.94 &  0.57/0.82 &   0.6/0.79 &  0.51/0.98 &  0.53/0.95 &  0.57/0.94 \\
\cline{2-8}
  & 0.8 & 0.83/0.94 &  0.83/0.86 &  0.83/0.82 &  0.81/0.98 &  0.82/0.97 &   0.8/0.95 \\
\hline
\end{tabular}
\caption{Experimental results of the four sampling algorithms evaluated with the two best performing GBM and LSTM models from Tables \ref{tab:2} and \ref{tab:3}. Precision/Recall scores are illustrated for h-coverages of 0.2, 0.5, 0.8 and document lengths $<$ 200 words, 200-500 words, and above 500 words. Test dataset described in Figure. \ref{fig7}. }
\label{tab:4}
\end{table*}

The most striking increase in model performance is given by the inclusion of punctuation in the input sequences. Without punctuation grammar is absent from text, and we theorized that keeping punctuation intact should increase model performance most significantly since it supports better understanding of context. We expected the performance boost to be significant only in the recurrent neural networks since this type of modeling has the ability to take advantage of sequential inputs. It is indeed the case - there was no improvement noticed for the GBM models - as even the worst punctuated models outperform the best of their counterparts. (see Table \ref{tab:3})

\subsection{Comparison of sampling algorithms}%Q1 demoted to subsection
Figure \ref{fig6s} shows the thorough evaluation of the best classifiers on the dataset in Fig. \ref{fig7}. The two parameters at play are H-coverage and document length. H-coverage, on the one hand, can be considered to be a latent variable, since we cannot know its value ahead of time. Document length, on the other hand, is known and can be used to make an informed decision about sampling algorithm choice prior to its use. 

One immediate trend that emerges is the superiority of the \textit{possum} method to all others in terms of sensitivity. The recall of this approach should be no surprise since this sampling algorithm considers all fragments with a score above 0. At the opposite end of the spectrum is the \textit{seq11} method (green line Figure \ref{fig6s}) which scores poorest. In all fairness to the algorithm, the model was trained on a distribution of samples, 97\% of which were more than one sentence long. \textbf{We can conclude that for a model with \textit{ high recall}, all ranges of h-coverage and document length \textit{possum sampling method} is superior. }

Another trend is that precision increases as documents get longer. This indicates that classifiers have good grasp of the concept of a highlight. Imagine hitting a target as the target gets larger in size. If you know where the center of the region with the highest score is located, as the target gets larger you will only hit this region more. Nevertheless, we point out that while it might appear that the \textit{seq11} model has superior precision to all the others, this is to be taken with a grain of salt (Table \ref{tab:4}). It's diagnostic ability is barely above that of chance level, as indicated in the ROC AUC plots (bottom). We can surmise that the \textit{seq11 sampling method} has the lowest predictive power. \textbf{The most reliable model with the highest precision, for all ranges of h-coverage and document length choose is then the hscore sampling method.}

Comparisons among the LSTM and GBM predictors are also subjective to h-coverage and document length. An average of all the sampling methods for LSTM models yields 0.583/0.554, while GBM 0.561/0.64 for precision/recall. GBM models perform better when it comes to recall, worse when it comes to precision. In terms of diagnostic ability, for high h-coverage the performance is similar, while for h-coverage of 0.2 - 0.5 we see a higher variance for the GBM models, with a peak in performance for videos of 30-40 minutes. It is interesting to observe that some models demonstrate linear behavior with respect to document length (\textit{hscore} ROC AUC in the LSTM at 0.5 h-coverage) while others are convex shaped. \textbf{4. Although LSTMs greatly outperform GBM in the highlight identification, the limitations of the sampling algorithm greatly diminish their performance.}

For these models to be implemented in a realistic production environment, one must also assess size of the models on disk as well as speed of performance. For example, stemming is a very time consuming task. New methods must be constantly evaluated and the additional accuracy gains need to be weighed against the engineering effort needed to bring them into a production environment. 

\begin{figure}[ht]
\centering
{
\includegraphics[width=1\linewidth]{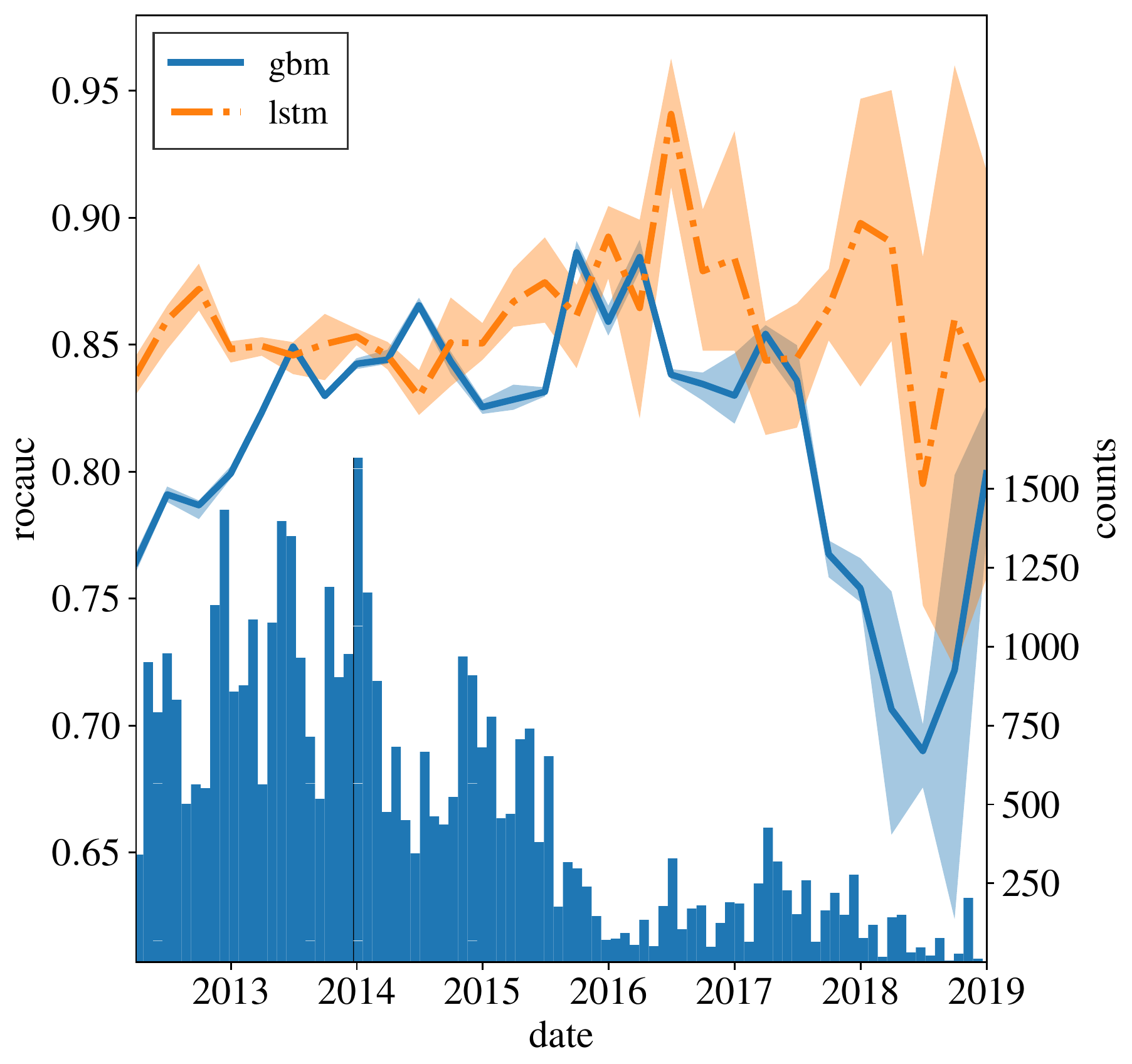}
}
\caption{Time dependent ROC AUC scores of the best LSTM/GBM classifiers shown as a function of the date of the test set samples. Best and worst scores are highlighted, solid line is the average performance. The distribution of training data is shown underneath. Performance of the models varies is significantly less in regions of abundant data and the vice versa. }
\label{fig8}
\end{figure}

\subsection{Time dependent covariate shift}
Over time, all systems evolve. External variables such as societal and regime changes, re-branding, new scientific discoveries and shifts in environment can all have an impact on market research methodologies. Thus, from time to time all models have to be re-trained and re-tested and possibly completely rebuilt. One excellent way to monitor results is to observe residual difference between forecasted and observed values. Such charts can be built into the software that implements the model itsel and the system monitors itself. Another method is to evaluate the variance in model performance as a function of time. 

We investigate the latter quantity. Figure \ref{fig8} shows the ROC AUC of the best LSTM and GBM models evaluated on test data from the past years. A flat line indicates the model has good robustness when it comes to time sensitive covariate shift, while regions of high variance implies one of two things. Either the data over the years does experience covariate shift or the sparsity of data in this region of time causes a performance dip. Unfortunately, we cannot control the quantity of data available.

If the performance dip is drastic, serious consideration should be given to re-training the model with a smaller dataset, spanning the years of interest. In this fashion, although the model will be lest robust, it will perform better locally and possibly in the near future. This assumes that the definition of a highlight is a local construct. From Figure \ref{fig8} we can see the LSTM demonstrates more stability in performance over the years even though its variance goes up significantly after 2017. The high variance exhibited in 2018 indicates less confidence in predictions of contemporary highlights.

\section{Conclusions and Future Work}
This study was inspired by a rising demand in the market research industry for technology to help researchers sift through an increasing supply of data. In this paper we have presented a novel approach for NLP based highlight identification and extraction based on a supervised learning approach. We also extracted, parsed and prepared a novel dataset for the task of highlight identification in market research transcripts. 

We constructed and evaluated multiple GBM architectures based on BOW/BONG feature extraction as well as LSTM methods sensitive to the intrinsic time ordering present in language. Models were perfected with only a single round of hyper-parameter tuning, and as such there is room for improvement. Nevertheless, we have achieved exceptionally well performing-models with ROC AUC scores in the range 0.93-0.94. We demonstrated that inclusion of punctuation into the input dataset and personalized embedding vectors are crucial for achieving a model with high diagnostic ability. Furthermore, we showed that incorporating an attention mechanism into the modeling, though it didn't improve performance, it dramatically reduced the training time by an almost an order of magnitude. 

Perhaps more importantly, we studied the effects of four different fragmenting and sampling methods on large text, prior to evaluating each fragment with an estimator of choice.  Overall we observed a decrease in the performance of the models, primarily due to the inefficient sampling choices. Nevertheless, we were able to determine some baseline values on which further work can improve upon, and extract some insights into algorithm choices. The \textit{possum} sampling method is ideal to use for high recall, and \textit{hscore} algorithm is preferred in terms of precision.

There is a lot of room for improvement in the fragmentation and sampling algorithms. The main issue is extreme redundancy in evaluating the elements of Eq. \ref{eq:frag_set}. One possibility is the design of intelligent sampling based on topic modeling. With such an approach, some redundancy might be avoided. Alternately, there are parallels between machine translation techniques which can be applied here, specifically that of beam search. There are additional architectures that we have not considered, but might be of interest for future work: i) single sentence highlight classifier, ii) auto-encoder and re-constructor approach and an approach taking advantage of recent developments in word embeddings - BERT Transformer \cite{bert}.

\section{Acknowledgments}
This work was supported by FocusVision Inc. (\textit{http://www.focusvision.com}). We are grateful to Tate Tunstall and Arunima Grover for discussions and suggestions and Rick Sullivan, Jake Kaad and Jonathan So for their technical assistance. 

\appendix

\section{Appendix}

\subsection{Generalized N-gram model} %Q1 moved from section "Model Design"

To construct a N-gram model, we generate tokens based on time ordered n-pairs of words. A document is then expressed as the linear combination 
\begin{eqnarray}
    \textbf{D}^{(M)} = \sum_{n=0}^M  \sum_i \alpha_n c_i^n T_i^{(n)}
\label{eqngram}
\end{eqnarray}
in orthogonal space spanned by $\mathcal{N}$ vectors $T_i^{(n)}$. Here $T_i^{(n)} = \{(w_i,w_{i+1}, ...w_{i+n}) | i+n < N^{(n)} \}$ where $N^{(n)}$ is the size of the vocabulary of the respective n-gram elements. The case n=0, corresponds to the uni-gram model where $N^{(0)}$ is the number of unique words in the corpora. If no other transformations are applied to vectors $T_i^{(n)}$ this is simply a bag of words (BOW) model.  This can be extended to show $\mathcal{N} = \sum_{n=0}^M \alpha_n N^{(n)}$. While coefficients $c_i$ are directly related to the semantic description of the document, and may be acted no by various rotation matrices, the $\alpha_n = {0,1}$ are manually set model hyperparameters: the set $\{M=1, \alpha_0 = 0, \alpha_1 = 1\}$  corresponds to bag of bi-grams model).

\subsection{Bag of N-grams}%Q1 moved from section "Model Design"

The BOW, and more generally, the bag of n-grams (BONG) models will serve as baselines. Albeit simple, there are many applications where their performance rivals that of sophisticated models with complex features representations \cite{bow1}.  All BONG/BOW models are non-Markovian and use the n-grams generated according to Eq. \ref{eqngram} as features, while ignoring order or relationships between neighbouring words. Semantic meaning is sometimes lost in models with n=1 due to existence of polysemous words - an ambiguity which higher n-gram models attempt to correct. 

Nevertheless, the intuition behind such models is that there is a unique set of words that can discriminate between document classes. This is obviously useful when trying to discern between classes such as "food" (some representative words being "ketchup", "fork", "oven") and "cars" (representative words being "breaks", "engine", "mpg"), but it isn't clear whether it will perform well for our somewhat ambiguous problem of discriminating between highlights and non-highlights. 

\subsection{Word and document embedding}%Q1 moved from section "Model Design"

A common alternate to representing documents as vectors described by Eq. \ref{eqngram} is to generate word embeddings. The advantage is the ability to capture similarity between words the model may have never seen. One hot word encoding can lead to dictionaries of very large dimension, typically sparse, and usually computationally expensive. Embeddings are a dense representation of words by a non-orthogonal set of latent vectors typically of much lower dimension \cite{embeddings1} than BONG/BOW models. Given a sentence with words $w_{ij}, t \in [0,T]$ we transform the words into vectors via an embedding matrix $W_{e}$ such that $v_{ij} = W_{e} w_{ij}$. The matrix $W_{e}$ can be trained in conjunction with the neural language model, however it is sometimes advantageous to use pre-trained weights. We will consider three cases: no embeddings, GloVe embeddings\cite{glove1}, and our own embeddings trained with AWS's BlazingText algorithm \cite{blazing_text}.  

For doc2vec embedding we use two schemes. Given $v_i \in \rm I\!R^{K}$, an embedding vector for word $w_i$, in document D the word occurs $n_i$ times. A generalized document embedding can be computed as 
\begin{equation}
    \textbf{L} = \sum_i c_i \textbf{tf}_i v_i
\label{eq:bom1}
\end{equation}
where $\textbf{tf}_i$ is the term frequency computed as $\textbf{tf}_i = n_i/N$, N being the total number of words in the document $N = \sum_i n_i$. In the simplest case we set $c_i = 1$ which is equivalent to a mean embedding model. We also build a TFIDF model which is an attempt at corrections for the large weights assigned to frequent words. Here we have set
\begin{equation}
    c_i = \textbf{idf}_i  = \log{\frac{1+D_N}{1+\textbf{df}_i}} + 1
\label{eq:bom2}
\end{equation}
where $D_N$ is the total number of documents in the corpus, and $\textbf{df}_i$ is number of documents in the document set that contain the word i. For out of dictionary (OOD) words, we assign them $max(\textbf{idf}_i)$ since it should be at least as infrequent as any of the known words. Although there are more sophisticated ways to approximate OOD words \cite{wordaprox} we set the $<UNK>$ word token to a mean embedding vector. 

\subsection{Long Short-term Memory with Attention (LSTM)}%Q1 moved from section "Model Design"

LSTM's are a type of feed-forward networks which attempt to solve the known gradient vanishing or exploding problems. \cite{lstm1} It has proven quite effective at handling a number of NLP tasks \cite{lstm2}. It models word sequences \textbf{w} as follows:
\begin{eqnarray}
    \textbf{i}_t &=& \sigma(\textbf{v}_t \textbf{U}^i + \textbf{h}_{t-1}\textbf{W}^i + \textbf{b}_i) \nonumber \\
    \textbf{f}_t &=& \sigma(\textbf{v}_t \textbf{U}^f + \textbf{h}_{t-1}\textbf{W}^f + \textbf{b}_f) \nonumber \\
    \textbf{o}_t &=& \sigma(\textbf{v}_t \textbf{U}^o + \textbf{h}_{t-1}\textbf{W}^o + \textbf{b}_o)  \\
    \textbf{q}_t &=& \tanh(\textbf{v}_t \textbf{U}^q + \textbf{h}_{t-1}\textbf{W}^q + \textbf{b}_q) \nonumber \\
    \textbf{p}_t &=& \textbf{f}_t \ast \textbf{p}_{t-1} + \textbf{i}_{t} \ast \textbf{q}_t \nonumber \\
    \textbf{h}_t &=& \textbf{o}_t \ast \tanh(\textbf{p}_t) \nonumber
\end{eqnarray}
where $i_t$, $f_t$ and $o_t \in \rm I\!R^{d x 2d}$ are the input, forget and output gate matrices, respectively, and $b_i$, $b_f$, $b_o \in \rm I\!R^{d}$ the biases. $\sigma$ i the sigmoid function, and $\ast$ stands for element wise multiplication. $v_t$ are the inputs to the LSTM, representing the word embedding vectors, and $h_t$ are the corresponding hidden states. All are parameters to be learned during training. 

Not all words are expected to contribute equally to the text evaluation. Hence, we introduce a hierarchical attention mechanism \cite{attention1} 
\begin{eqnarray}
    \textbf{u}_{t} &=& \tanh(\textbf{h}_t \textbf{U}^w + \textbf{b}_w) \nonumber \\
    \bm{\alpha}_{t} &=& \frac{\exp(\textbf{u}_{t}^\top \textbf{u}_{w})}{\sum_t \exp(\textbf{u}_{t}^\top \textbf{u}_{w})} \\ 
    \textbf{c} &=& \sum_t \bm{\alpha}_{t} \textbf{h}_t \nonumber
\end{eqnarray}
The hidden vectors $\textbf{h}_t$ are fed through a secondary feed-forward neural network which is jointly trained with the LSTM to obtain $\textbf{u}_t$ as the hidden representation. To compute word importance vectors $\bm{\alpha}_{t}$ we compute the normalized similarity of $\textbf{u}_t$ with some word level context vector $\textbf{u}_w$. The context vector is a key component, as it can be seen as a high level representation of "informative words" in memory networks. Instead of using the final hidden state to compute a probability, as is typical in LSTMs, we now use the context vector \textbf{c} which is a weighted sum over all of the hidden states. 
\begin{equation}
    p = softmax(\textbf{c} \textbf{W}_c + \textbf{b}_c) 
\end{equation}
In this architecture the probability $\bm{\alpha}_{t}$ reflects the importance of hidden vector $\textbf{h}_t$. Intuitively this corresponds to the amount of "attention" the estimator is paying to a given word. 

\bigskip

\bibliography{references}
\bibliographystyle{aaai}

\end{document}